\documentclass{article} 
\usepackage{iclr2022_conference,times}

\usepackage{hyperref}
\usepackage{url}

\usepackage{amsmath}
\usepackage{amsfonts}

\usepackage{algorithm}
\usepackage{algpseudocode}
\usepackage{graphicx}
\usepackage{subfigure}
\usepackage{color}
\usepackage{bm}
\usepackage{multirow}
\usepackage{booktabs}       

\title{Off-policy Imitation Learning from Visual Inputs}

\iclrfinalcopy

\author{Zhihao Cheng \\
The University of Sydney\\
\texttt{zche3121@uni.sydney.edu.au}
\And
Li Shen \\
JD Explore Academy \\
\texttt{mathshenli@gmail.com} \\
\AND
Dacheng Tao \\
JD Explore Academy \& The University of Sydney \\
\texttt{dacheng.tao@gmail.com}
}

\definecolor{target_q_aug}{rgb}{0.43921568627,0.67843137254,0.27843137254}
\definecolor{q_aug}{rgb}{0.26666666666,0.44705882352,0.76862745098}
\definecolor{f_func}{rgb}{1.0, 0.49019607843, 0.19215686274}

\begin{document}

\maketitle

\begin{abstract}
  Recently, various successful applications utilizing expert states in imitation learning (IL) have been witnessed. However, another IL setting---IL from visual inputs (ILfVI), which has a greater promise to be applied in reality by utilizing online visual resources, suffers from low data-efficiency and poor performance resulted from an on-policy learning manner and high-dimensional visual inputs. We propose OPIfVI (Off-Policy Imitation from Visual Inputs), which is composed of an off-policy learning manner, data augmentation, and encoder techniques, to tackle the mentioned challenges, respectively. More specifically, to improve data-efficiency, OPIfVI conducts IL in an off-policy manner, with which sampled data can be used multiple times. In addition, we enhance the stability of OPIfVI with spectral normalization to mitigate the side-effect of off-policy training. The core factor, contributing to the poor performance of ILfVI, that we think is the agent could not extract meaningful features from visual inputs. Hence, OPIfVI employs data augmentation from computer vision to help train encoders that can better extract features from visual inputs. In addition, a specific structure of gradient backpropagation for the encoder is designed to stabilize the encoder training. At last, we demonstrate that OPIfVI is able to achieve expert-level performance and outperform existing baselines no matter visual demonstrations or visual observations are provided via extensive experiments using DeepMind Control Suite.
\end{abstract}

\section{Introduction}
Imitation learning (IL) empowers agents to learn from expert data instead of designing an explicit reward function \citep{ho2016generative} and has achieved remarkable successes in graphics \citep{yuan20183d}, online games \citep{vinyals2019grandmaster}, robotic manipulation \citep{fang2019survey}, and saliency prediction \citep{xu2021saliency}. The expert data that IL uses can be divided into two categories \citep{goo2019one,wake2020learning}, demonstrations and observations, among which demonstrations contain states and actions of experts' experiences, whereas observations only consist of states. In real world applications, 
the state is the proprioceptive state of an expert, which could be hard to access and record. By contrast, intelligent creatures grasp knowledge or skills by observing how peer fellows accomplish tasks without knowing their proprioceptive states~\citep{douglas2006observational}. In other words, intelligent creatures generally learn with visual inputs rather than state inputs. This learning scheme is more practical but has been less studied in IL community.

The thought to enable agents to learn like intelligent creatures lead to a spectrum of IL, imitation learning from visual inputs (or images, or pixels) (ILfVI). This spectrum of IL is also referred to as visual imitation learning (VIL) \citep{young2020visual,rafailov2021visual}. Here, we denote it as ILfVI to emphasize that visual inputs can be further classified. Corresponding to traditional state demonstrations and observations, visual inputs fall into visual demonstrations and visual observations, where the former contains images and actions while the latter merely includes images. In contrast to dramatic successes in IL from state inputs \citep{ho2016generative,torabi2018generative, zhang2020f}, there is only a few research \citep{li2017infogail,torabi2018generative} on ILfVI. Furthermore, the performance of ILfVI is still far from satisfactory to be applied in reality.

Compared to state inputs, the only difference in ILfVI is that images are partially-observed high-dimensional inputs. This difference introduces several challenges: 1) visual inputs are partially observed from states, which converts underlying dynamics from Markov Decision Process (MDP) \citep{bellman1957markovian} into Partially Observable MDP (POMDP) \citep{kaelbling1998planning}; 2) high-dimensional inputs such as images contain a large portion of redundant information \citep{ramachandran2019stand}, distracting agents from extracting useful information for decision-making; 3) the adversarial training paradigm in IL suffers from training instability with high-dimensional inputs \citep{goodfellow2014generative}. In addition, agents in ILfVI need to learn to extract meaningful features from high-dimensional inputs, further aggravating low data-efficiency in an on-policy training manner. Despite of these challenges, it is significant to improve the performance of ILfVI because this IL setting extremely expands promising application domains.

In this paper, considering the challenges in ILfVI, we propose the algorithm OPIfVI (Off-policy Imitation from Visual Inputs) to improve both data-efficiency and performance. To be more data-efficient, we build OPIfVI in an off-policy manner, where sampled data are stored in a replay buffer such that the data can be utilized multiple times. We adopt spectral normalization further to enhance the stability of this off-policy training scheme. Besides, we borrow the idea of data augmentation from computer vision to help encoders to extract meaningful features from images. Then, the extracted features are forwarded to agents to make decisions. Data augmentation enlarges sampled data and can benefit data-efficiency to some extent. Furthermore, we design a specific structure for the gradient backpropagation to train and stabilize encoders. In this structure, the actor and critic in the generator share an encoder, while the discriminator maintains an independent encoder. The encoder in the generator is updated with only gradients backpropagated from the critic, whereas the other encoder is trained with the discriminator loss. Combing these three aspects, we propose the algorithm OPIfVI, which can efficiently and effectively learn from visual inputs. We evaluate OPIfVI's ability to reproduce expert policies with visual demonstrations and visual observations on various DeepMind Control tasks \citep{tassa2018deepmind}, showing that OPIfVI significantly surpasses the other baselines in terms of both data-efficiency and performance. Finally, ablation studies are carried out to investigate the impacts of modifications that we adopt.

The reminder of this work is organized as follows. We present the related work in Section \ref{sec:related_work} and introduce necessary background knowledge in Section \ref{sec:preliminary}. Then, our method OPIfVI is thoroughly described in Section \ref{sec:method}. Section \ref{sec:experiments} empirically demonstrates the performance of OPIfVI, and Section~\ref{sec:conclusion} concludes the paper.

\section{Related Work}\label{sec:related_work}
\paragraph{Imitation Learning from State Inputs.} Imitation Learning (IL) allows reproducing policies that can imitate expert behaviors merely relying on expert data. IL algorithms can be split into different classes from various perspectives. For example, according to the learning mechanism, IL can be divided into behavioral cloning (BC) \citep{Bain95} and inverse reinforcement learning (IRL) \citep{abbeel2004apprenticeship}. BC takes IL as pure supervised learning, while IRL first reconstructs a reward function with expert data and then conducts ordinary RL with the reconstructed reward function. For more works, please refer to \citet{ho2016generative,fu2017learning,torabi2018generative, zhang2020f, dadashi2020primal, jaegle2021imitation} and their references therein. From the perspective of expert data, IL can be categorized into learning from demonstration (LfD) and learning from observation (LfO) \citep{goo2019one,wake2020learning}. Demonstrations contain both states and actions of experts, whereas observations only consist of expert states. \citet{torabi2018behavioral} present behavioral cloning from observation (BCO) by combining BC with an inverse dynamics model to forecast expert actions. \citet{torabi2018generative} develop generative adversarial imitation from observation (GAIfO) that utilizes state transitions rather than state-action pairs to generate rewards, extending GAIL to the LfO setting. Most of these IRL algorithms employ an on-policy learning manner to maintain accurate estimations of occupancy measures \citep{ho2016generative,torabi2018generative}, which results in low data-efficiency.

\paragraph{Imitation Learning from Visual Inputs.}
ILfVI is attracting more attention owing to its broad application prospects. However, there is only a few research on ILfVI. \citet{li2017infogail} propose InfoGAIL that can deal with visual demonstrations sampled from diverse experts by learning latent variables from expert data. To cope with visual demonstrations, \citet{young2020visual} enhance BC with data augmentation and develop visual behavior cloning (VBC). \citet{torabi2018generative} conduct experiments of GAIfO with visual observations, showing that GAIfO only achieves about half of the expert-level performance in no-trivial environments. Two concurrent works \citep{rafailov2021visual,anonymous2022imitation} give further insights into ILfVI. \citet{rafailov2021visual} solve ILfVI from a model-based perspective, whose algorithm V-MAIL first learns a world model and then updates the discriminator with on-policy samples from the learned model. \citet{anonymous2022imitation} study IL from visual observations with a model-free scheme. They build their IL algorithms based on the encoder in DrQ-v2 \citep{yarats2021mastering} to help extract features and achieve expert-level performance. Another spectrum of researchers focus on ILfVI in the domain of robot control \citep{yu2018one,pathak2018zero}. They aim to achieve few-shot or even zero-shot IL, and expert data are provided with time labels, which is distinct from our setting. 

\paragraph{Data Augmentation.} In computer vision (CV), data augmentation is one of the basic techniques for the majority of tasks such as classification \citep{perez2017effectiveness}, detection \citep{zoph2020learning}, and recognition \citep{lv2017data}. Data augmentation has long been studied in CV~\citep{fawzi2016adaptive,cubuk2019autoaugment,tran2021data}, which helps enlarge dataset and extract useful features. However, for state inputs, data augmentation is rarely adopted in RL or IL \citep{yarats2020image} because the state $s$ in MDP is unique and any modification will lead to a state that represents different information compared to the unmodified one. \citet{sinha2021s4rl} study how to augment states in RL by adding small noises to sampled data. For visual inputs, recently, data augmentation has been employed in RL and significantly improves the performance \citep{yarats2019improving,yarats2020image,laskin2020reinforcement,seo_chen2021re3}. For example, \citet{laskin2020reinforcement} illustrate that general data augmentation methods enable agents to achieve excellent performance in RL from visual inputs with extensive studies. In ILfVI, data augmentation also demonstrates remarkable performance gains \citep{young2020visual,rafailov2021visual}. 

Our work bears some resemblance to the two concurrent works \citet{rafailov2021visual,anonymous2022imitation}. Compared to \citet{rafailov2021visual}, we solve ILfVI from a model-free perspective, which improves the performance with merely off-policy samples instead of a learned model and even works for visual observations. In contrast to \citet{anonymous2022imitation}, we dynamically calculate rewards with the latest discriminator and employ spectral normalization to stabilize the learning process, which surpasses the former in terms of both data-efficiency and performance. 

\section{Preliminaries}\label{sec:preliminary}
\paragraph{Markov Decision Process (MDP) \citep{bellman1957markovian}.} We consider an MDP described by a tuple $(\mathcal{S},\mathcal{A},\mathcal{T},r,\gamma)$, with the state space $\mathcal{S}$, the action space $\mathcal{A}$, the transition distribution $\mathcal{T}=\mathcal{T}(s'|s,a)$, the reward function $r:\mathcal{S} \times \mathcal{A} \rightarrow \mathbb{R}$, and the discount factor $\gamma$. We denote a stochastic policy for the agent as $\pi(a|s):\mathcal{S}\times \mathcal{A} \rightarrow[0,1]$, where $s\in \mathcal{S}$ and $a \in \mathcal{A}$. A trajectory $\tau =\{s_t,a_t\}_{t=0}^{\infty}$ can be obtained via interactions between policy $\pi(a|s)$ and the environment, where $t$ is the current timestep, the initial state $s_0$ is sampled from the probability distribution $s_0\sim\rho_0(s_0)$, $a_t\sim\pi(a_t|s_t)$, and $s_{t+1}\sim \mathcal{T}(s_{t+1}|s_t,a_t)$. Then, the expected discounted reward is $J(\pi)=\mathbb{E}_{\tau \sim \pi}\left[\sum_{t=0}^{\infty}\gamma^t r(s_t,a_t)\right]$. RL algorithms are supposed to find the optimal policy $\pi^{*}(a|s)$, which can achieve the maximum episode cumulative reward $J^*(\pi)$.

When we use visual inputs to control agents, the MDP formulation turns into POMDP \citep{yarats2020image}. Compared to MDP, POMDP can be formulated with a 7-tuple $(\mathcal{S},\mathcal{A},\mathcal{T},r,\gamma, \Omega, O)$, where the two additional elements $\Omega$ and $O$ is the space of observations and the probability that $O(o_{t+1}|s_t,a_t)$, respectively. Agents can merely receive partially observed information $o$ instead of state $s$ in POMDP, increasing the difficulty of making decisions. A routine solution to deal with the partial observability is to stack several adjacent visual inputs together and then regard it as a state $s_t\approx \bm{o_t}=\{o_t,o_{t-1},\cdots \}$ \citep{mnih2015human,laskin2020reinforcement,yarats2020image}. In the paper, we follow this routine to cope with visual inputs and then carry out IL.

\paragraph{Soft Actor-Critic (SAC) \citep{haarnoja2018soft}.} We introduce the off-policy RL algorithm SAC, which maintains a policy $\pi_\theta$, two Q-value functions $Q_{\phi_1}$ and $Q_{\phi_2}$, and a temperature parameter $\alpha$. SAC uses samples from a replay buffer $\mathcal{R}$ and updates the Q-value functions to minimize the following objectives
\begin{align}
J_Q(\phi_i)=\mathop{\mathbb{E}}\limits_{(s,a,r,s',d)\sim \mathcal{R}} \left( Q_{\phi_i}(s,a)-(r+\gamma(1-d)y) \right)^2,
\end{align}
where $i\in\{1,2\}$, $d$ is the done signal ($d=1$ if an episode terminates; otherwise $d=0$) and $y=(\mathop{\min}\limits_{i=1,2}Q_{\bar \phi_i}(s',a')-\alpha \log \pi_\theta(a'|s'))$. $Q_{\bar \phi_i}$ is defined as the target Q-value regarding $Q_{\phi_{i}}$ and its parameters $\bar \phi_i$ can be obtained via exponential moving average of $\phi_{i}$. The policy parameters $\theta$ are updated to maximize
\begin{align}
    J_{\pi}(\theta)=\mathop{\mathbb{E}}\limits_{s\sim \mathcal{R}, a\sim \pi_\theta(a|s)} \left(\mathop{\min}\limits_{i=1,2}Q_{\phi_{i}}(s,a)-\alpha \log \pi_\theta(a|s)\right).
\end{align}
The temperature $\alpha$ that helps improve stability is adjusted by minimizing
\begin{align}
    J_{\alpha}(\alpha) = \mathop{\mathbb{E}}\limits_{s\sim \mathcal{R}, a\sim \pi_\theta(a|s)} \left(-\alpha \log \pi_\theta(a|s)-\alpha \mathcal{\bar H} \right),
\end{align}
where $\mathcal{\bar H}$ is the target minimum entropy.

\paragraph{Generative Adversarial Imitation Learning (GAIL) \citep{ho2016generative}.} GAIL adopts the framework of GAN (Generative Adversarial Network) training \citep{goodfellow2014generative} and can be formalized as a minimax problem: 
\begin{align}
    \begin{split}
        \mathop{\min}\limits_{\theta} \mathop{\max}\limits_{\omega} \; \mathbb{E}_{\pi_\theta}[\log D_\omega(s,a)]+\mathbb{E}_{\pi_E}[\log(1- D_\omega(s,a))]-\beta \mathcal{H}(\pi_{\theta}),
    \end{split}
\end{align}
in which $D_\omega(s,a)$ is the discriminator that measures the similarity between agent's state-action pair $(s,a)$ and the expert one, $\omega$ is the parameter of the discriminator, and $\mathcal{H}(\pi)=\mathbb{E}_{\pi}[-\log \pi(a|s)]$ is the entropy of policy $\pi$ with weight $\beta \ge 0$.

\paragraph{Generative Adversarial Imitation from Observation (GAIfO) \citep{torabi2018generative}.} GAIfO is extended from GAIL to learn from expert data with the absence of actions. The only difference lying between the two approaches is that GAIfO uses state-state $(s,s')$ pairs, whereas GAIL utilizes state-action $(s,a)$ pairs. GAIfO is formalized as follows
\begin{align}
    \begin{split}
        \mathop{\min}\limits_{\theta} \mathop{\max}\limits_{\omega} \; \mathbb{E}_{\pi_\theta}[\log D_\omega(s,s')]+\mathbb{E}_{\pi_E}[\log(1- D_\omega(s,s'))].
    \end{split}
\end{align}
When using single states instead of state-state pairs, GAIfO degrades to GAIfO-s as follows \citep{yang2019imitation},
\begin{align}
    \begin{split}
        \mathop{\min}\limits_{\theta} \mathop{\max}\limits_{\omega} \; \mathbb{E}_{\pi_\theta}[\log D_\omega(s)]+\mathbb{E}_{\pi_E}[\log(1- D_\omega(s))].
    \end{split}
\end{align}

\section{Off-policy Imitation from Visual Inputs}\label{sec:method}

\begin{figure*}[!htb]
  \centering
      \includegraphics[width=0.7\textwidth]{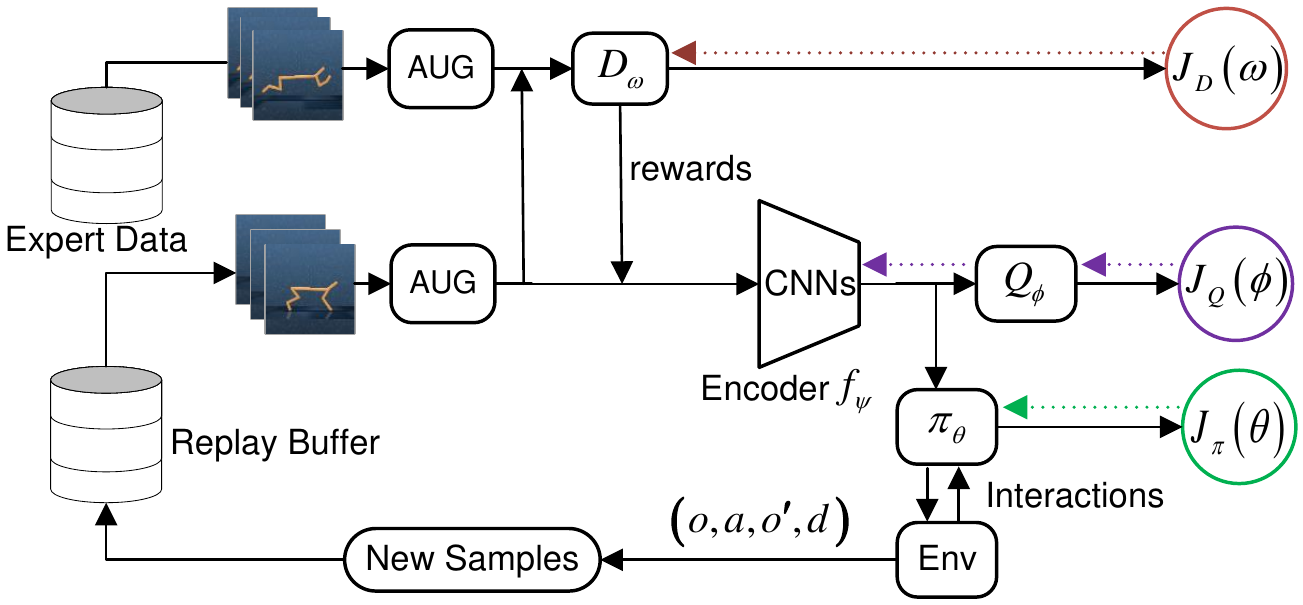}
  \caption{The framework of OPIfVI. The main components in OPIfVI are a replay buffer, an expert data set, a data augmentation block (abbreviated as \footnotesize{AUG}), an encoder $f_\psi$, a policy $\pi_\theta$, the Q-value function $Q_{\phi}$, and a discriminator $D_\omega$. The solid lines with arrows denote directions of information flow, while the colored dotted lines represent gradients backpropagated from loss functions.}
  \label{fig:framework}
\end{figure*}

Here, we describe our off-policy imitation algorithm for learning from visual inputs, OPIfVI, which is presented in Figure \ref{fig:framework}. We begin by formalizing the off-policy ILfVI problem and introducing the challenges responsible for low data-efficiency and poor performance of current IL algorithms. To alleviate these challenges, OPIfVI adopts three major modifications compared to previous works: 1) an off-policy imitation learning paradigm with enhanced stability (Section \ref{subsection:off-policy}); 2) data augmentation for better feature extraction (Section \ref{subsection:dataaugment}); 3) a specifically designed gradients backpropagation scheme for the training of encoders (Section \ref{subsection:encoder}). By virtue of the three modifications, our algorithm--OPIfVI is able to primely achieve expert-level performance in ILfVI with high data-efficiency, which surpasses the other baselines. 

\paragraph{Problem Formulation.} Two major differences are lying between our ILfVI setting and the previous ones \citep{ho2016generative,torabi2018generative}. The first is that agents only receive high-dimensional partially-observed images instead of low-dimensional fully-observed states. Although we can stack several consecutive images into $\bm{o}$ and roughly regard $\bm{o}$ as a state, the high-dimensional inputs still make it challenging to imitate. We consider the problem of learning from visual demonstrations and observations, \emph{i.e.}, the expert data are $\tau_E=\{(\bm{o},a)\}_0^N$ or $\tau_E=\{(\bm{o},\bm{o'})\}_0^N$, respectively. Furthermore, we also study a degraded setting in visual observations, where only single observations rather than neighboring observation pairs are provided such that $\tau_E=\{(\bm{o})\}_0^N$. To unify the three kinds of expert data, we define a symbol $\bm{x}$ such that $\bm{x}=(\bm{o},a)$, $(\bm{o},\bm{o'})$, or $(\bm{o})$. The remaining difference is that we use off-policy samples to conduct IL. With off-policy samples, the source distribution of samples for updating parameters would substantially change, increasing training instability. Our ILfVI problem is formalized as follows:
\begin{align}
    \begin{split}
        \mathop{\min}\limits_{\theta} \mathop{\max}\limits_{\omega} \; \mathbb{E}_{\mu^\mathcal{R}(\bm{x})}[\log D_\omega(\bm{x})]+\mathbb{E}_{\pi_E}[\log(1- D_\omega(\bm{x}))],
    \end{split}
\end{align}
where $\mu^\mathcal{R}(\bm{x})$ is the distribution of $\bm{x}$ in a replay buffer $\mathcal{R}$. Samples in the replay buffer are recorded from historical policies, which is considered off-policy. 

\paragraph{Algorithm Overview.} The framework of OPIfVI is presented in Figure \ref{fig:framework}. Similar to SAC, there are four Q-value functions in OPIfVI, including two alternate Q-value functions ($Q_{\phi_1}$ and $Q_{\phi_2}$) and two target ones ($Q_{\bar \phi_1}$ and $Q_{\bar \phi_2}$). For simplicity, we denote them with a unified symbol $Q_\phi$ in Figure~\ref{fig:framework}. The policy $\pi_\theta$ and Q-value functions share an identical encoder $f_\psi$ that is constructed with convolutional neural networks (CNNs) and is used to extract features from images. With interactions between the policy $\pi_\theta$ and the environment, we collect samples and store them into a replay buffer $\mathcal{R}$. For training, we first randomly sample data from the replay buffer and augment sampled images with random crop. Then the augmented data are used to calculate rewards with the current discriminator $D_\omega$. Subsequently, we input the data into the encoder $f_\psi$ followed by the Q-value function $Q_\phi$ as well as the policy $\pi_\theta$. According to the losses in Algorithm \ref{algo:OPIfVI}, we backpropagate gradients to update the parameters $\phi$, $\theta$, $\psi$. Note that the encoder is only trained with gradients from Q-value functions. The discriminator $D_\omega$ maintains a separate encoder with the same structure to the one of the policy and Q-value functions. The discriminator uses samples from the replay buffer and expert data set to update its parameters $\omega$ with the loss in Algorithm \ref{algo:OPIfVI}. Images input to the discriminator are also augmented and spectral normalization is adopted to enhance the stability of discriminator training.

\begin{algorithm}[tb]
  \caption{Off-policy Imitation from Visual Inputs (OPIfVI)}
   \label{algo:OPIfVI}
\begin{algorithmic}
  \State {\bfseries Inputs:} Expert trajectories $\tau_E$.
  \State {\bfseries Hyperparameters:} Total iteration number $M$, replay buffer size $H$, initial number of samples $B$, image augmentation $\text{\footnotesize AUG}$, minibatch size $N$, learning rate $\eta$, polyak averaging for target networks $\rho$, discount factor $\gamma$, target minimum entropy $\mathcal{\bar H}$, and temperature $\alpha$.
  \State {\bfseries Parameters:} Denote the encoder as $f_\psi$, policy as $\pi_{\theta}$, Q-values and target Q-values as $Q_{\phi_i}$ and $Q_{\bar \phi_i}$ ($i\in\{1,2\}$), discriminator $D_{\omega}$. The parameters of each block are denoted by its subscript.
  \State {\bfseries Initialize Replay Buffer $\mathcal{R}$}
  \Comment Randomly sample $B$ transitions $(\bm{o},a,\bm{o}',d)$
  \For{$i=1$ {\bfseries to} $M$}
      \State $\{(\bm{o}, a, \bm{o}',d)\}_{k=1}^{N} \sim \mathcal{R}$ 
      \Comment Sample $N$ transitions from replay buffer $\mathcal{R}$
      \State $\{(\bm{o}, a, r, \bm{o}',d)\}_{k=1}^{N}$ with $r=-\log(D_{w}(\bm{o},a))$
      \Comment Compute rewards with discriminator $D_{w}$
      \State $\{(\bm{o}_e, a_e)\}_{k=1}^{N} \sim \tau_E$ \Comment Sample expert data
      \State $x=\text{\footnotesize AUG}(x), x\in \{\bm{o},\bm{o}',\bm{o}_e\}$
      \Comment Data augmentation
      \State $z=f_\psi(\bm{o}),z'=f_\psi(\bm{o}')$
      \Comment Extract features with the encoder
      \State  $y(r,z',d)=r+\gamma(1-d)\left(\mathop{\min}\limits_{i=1,2}Q_{\bar \phi_i}(z',a')-\alpha \log \pi_{\theta}(a'|z')\right), \; a' \sim \pi_{\theta} (\cdot |z')$
      \State  $\nabla_{\phi_i,\psi} \frac{1}{N} (Q_{\phi_i}(z,a)-y(r,z',d))$
      \Comment Update encoder $f_\psi$ and Q-value $Q_{\phi}$
      \State $Q_{\bar \phi_i} \leftarrow \rho Q_{\bar \phi_i} + (1-\rho) Q_{\phi_i}, \; i\in \{1,2\}$
      \Comment Update target Q-value functions
      \State $\nabla_{\theta} \frac{1}{N} \left(\mathop{\min}\limits_{i=1,2}Q_{\phi_{i}}(z,\bar a)-\alpha \log \pi_{\theta}(\bar a|z)\right), \; \bar a \sim \pi_{\theta}(\cdot|z)$
      \Comment Update policy $\pi_{\theta}$
      \State $\nabla_{\alpha} \frac{1}{N} \alpha (-\log \pi_\theta(\bar a|s)- \mathcal{\bar H}), \; \bar a \sim \pi_{\theta}(\cdot|z)$
      \Comment Update temperature $\alpha$
      \State $\nabla_\omega \frac{1}{N} \mathbb{E}_{(\bm{o}, a)}[ \log D_\omega(\bm{o},a)]+ \mathbb{E}_{(\bm{o}_e, a_e)}[\log(1- D_\omega(\bm{o}_e,a_e))]$
      \Comment Update discriminator $D_\omega$
      \State $\mathcal{R} \leftarrow \mathcal{R} \cup( \bm{o}, a, \bm{o}',d)$
      \Comment Sample a new transition to replay buffer $\mathcal{R}$
  \EndFor
\end{algorithmic}
\end{algorithm}

\subsection{Off-policy Learning}\label{subsection:off-policy}
To improve data-efficiency, OPIfVI adopts an off-policy training manner via an off-policy generator SAC. SAC uses a replay buffer $\mathcal{R}$ to store historical samples that are collected by previous policies and randomly samples data from this buffer to train its policy and Q-value networks \citep{haarnoja2018soft}. Such a replay buffer enables samples to be utilized multiple times for policy improvement, thus making learning more data-efficient. However, this off-policy learning scheme poses a threat to the training stability of OPIfVI.

Compared to on-policy adversarial IL algorithms \citep{torabi2018generative,zhang2020f}, OPIfVI updates the discriminator with data from a replay buffer to improve its ability on discriminating samples, resulting in an off-policy training mode for the discriminator. It is difficult to estimate the characteristics of the current generator and discriminator from off-policy samples as accurately as on-policy samples. As a result, the off-policy update of the adversarial training structure would be less stable \citep{kostrikov2019imitation}. What's worse, this off-policy regime is likely to over-fit to training data, leading to severe training instability or even failures of imitation as shown in \citet{rafailov2021visual,hoshino2021opirl}. In OPIfVI, to enhance the training stability against the drawbacks of off-policy learning, we employ spectral normalization \citep{miyato2018spectral,cheng2021guaranteed} to force the discriminator to be local Lipschitz-continuous. Local Lipschitz continuity of the learned reward function is necessary to achieve excellent performance of off-policy adversarial IL algorithms \citep{blonde2020lipschitzness}. 

With spectral normalization, the performance of OPIfVI as well as its stability are significantly improved. In particular, different from \citet{anonymous2022imitation}, OPIfVI does not save rewards into the replay buffer $\mathcal{R}$ and re-calculates rewards with the newest discriminator $D_\omega$ for the policy improvement. This dynamical reward calculation approach provides more accurate rewards for every update and is able to learn much faster than \citet{anonymous2022imitation}. 

\subsection{Data Augmentation}\label{subsection:dataaugment}
Unlike state inputs, where every element of a state $s$ stands for practical physical meanings and is irreplaceable, images contain plenty of redundant information. In ILfVI, agents struggle to select actions based on these high-dimensional inputs and need first to extract meaningful features from pixels. For example, in robot locomotion tasks, we suppose agents can accurately estimate the angles and angle velocities of robot joints, which is defined as the state of a robot \citep{peng2018deepmimic}, from images $\bm{o_t}$ for decision-making. However, it is challenging to learn what is essential for making decisions from several consecutive images. To help extract meaningful features from images, we employ data augmentation in OPIfVI. 

Data augmentation is used to enlarge expert data and agent data, and helps suppress overfitting as well as enhance robustness \citep{shorten2019survey}. In a way, the data-efficiency in OPIfVI is also improved because we can obtain more samples by augmenting $\bm{o_t}$. Due to the property of images, modifying a series of pixels in an image would not distort the core information \citep{shorten2019survey}. Hence, data augmentation helps OPIfVI to learn invariant features from images. In OPIfVI, we employ random crop to augment visual inputs, which is considered to be simple and can dramatically improve the performance \citep{yarats2020image}. Data augmentation is vital for the successful imitation of our algorithm OPIfVI, which is empirically studied in Subsection~\ref{subsection:ablationstudy}.

\subsection{Encoder Training Structure}\label{subsection:encoder}
As discussed in Subsection \ref{subsection:dataaugment}, images contain a large portion of redundant information that is useless for agents to make decisions. It is significant for ILfVI to extract useful features from images and prevent distracting agents from selecting proper actions by the useless information. Therefore, OPIfVI employs encoders to extract features from images. An encoder perceives an image in RGB form and outputs low-dimensional features \citep{wang2019discerning}. In our framework, three networks (the policy $\pi_\theta$, Q-value function $Q_\phi$, and discriminator $D_\omega$) need to use the encoder structure to extract features. As a result, how to regulate the encoders of those networks and properly train encoders to improve their ability on extracting features is challenging.

First, we share the encoder between the actor $\pi_\theta$ and the critic $Q_\phi$. We denote the encoder with $f_\psi$, which can output a latent feature from augmented adjacent images, $z_t=f_\psi(\text{\footnotesize AUG}(\bm{o_t}))$. After encoding, latent features are input to two separate MLPs (multilayer perceptrons) to build the policy and Q-value functions, leading to policy $\pi(a|\bm{o_t})=\pi_\theta(a|z_t)$ and Q-value $Q(\bm{o_t},a)=Q_\phi(z_t,a)$. The encoder's parameters are only updated with gradients from Q-value with losses in Algorithm \ref{algo:OPIfVI}. This separate update law is inspired by SAC-AE \citep{yarats2019improving}, which shows that training the encoder with only gradients from Q-value network performs better and more stable than training with the gradients from both the actor and critic. Second, the discriminator $D_\omega$ also needs an encoder block. The problem becomes how to design the encoder for the discriminator network.

Generally, we have three choices: 1) maintain a separate encoder for the discriminator; 2) share the encoder of Q-value functions with the discriminator and co-train the encoder with gradients from both the discriminator and Q-value functions; 3) share the encoder with the discriminator but do not backpropagate gradients from the discriminator to train the encoder. Resembling previous work GAIfO \citep{torabi2018behavioral}, we choose to hold an independent encoder for the discriminator rather than sharing the encoder of Q-value functions. We choose this structure due to the following reasons. The crucial role of a discriminator is to discriminate whether an image is sampled from the agent or the expert. Many successful GAN methods, whose discriminators possess a separate encoder, have been developed \citep{skorokhodov2021adversarial,karras2021alias}. Besides, our discriminator is spectral normalized, which enforces the Lipshitz continuity of networks. The Lipshitz continuity could impair the representational capacity of neural networks, making it difficult to share parameters of encoders between the discriminator and the generator. Furthermore, the discriminator $D_\omega$ is trained with losses such as binary cross entropy loss (BCE Loss \citep{osa2018algorithmic}). This kind of loss could distract the encoder from extracting meaningful features for decision-making.

In OPIfVI, we share an encoder between the policy and the Q-value network but maintain a separate encoder for the discriminator. The encoders help extract features from images for downstream applications, such as selecting actions or discriminating samples. The shared encoder between $\pi_\theta$ and $Q_\phi$ is trained with mere gradients from Q-value loss functions, while the encoder of the discriminator is updated with the discriminator loss. This design plays an important role in stabilizing the training of the imitator and achieves better performance compared to previous structures.

\section{Experimental Results}\label{sec:experiments}
\label{sec:result}
We conduct experiments with DeepMind Control Suite \citep{tassa2018deepmind} to demonstrate the performance of OPIfVI and compare it against other baselines. We aim to answer the following questions: 
\begin{itemize}
  \item [1)] 
  Can OPIfVI successfully reproduce expert policies from visual inputs and outperform other baselines regarding both data-efficiency and performance?  
  \item [2)]
  Does every modification that we adopt help improve the performance or data-efficiency of OPIfVI, and what role does it play. In particular, we investigate the effects of spectral normalization in the off-policy training manner, data augmentation, and the encoder training structure.
\end{itemize}

\subsection{Setups and Baselines}
We choose four typical environments in DeepMind Control Suite \citep{tassa2018deepmind}, \emph{i.e.}, CartPole Swingup, Walker Walk, Hopper Stand, and Cheetah Run. First, we use DrQ \citep{yarats2020image} to train experts in these environments and then sample data using trained experts. Visual observations can be obtained by removing actions in visual demonstrations. Then, we can conduct IL experiments with the acquired visual inputs. More details on environments, constructing expert data, and hyperparameters are deferred to the Appendix.  

We compare OPIfVI against two spectra of IL algorithms because visual inputs can be divided into visual demonstrations and visual observations. Hence, different baselines are employed corresponding to visual demonstrations or observations, which are briefly introduced below. To achieve fair evaluations, we use the identical data augmentation technique as in \citet{yarats2020image} and neural network architectures for inchoate algorithms.

\paragraph{Baselines for visual demonstrations} Corresponding to BC and IRL in IL, we select two baselines for visual demonstrations, \emph{i.e.}, VBC \citep{young2020visual} and P-DAC \citep{anonymous2022imitation}. VBC directly extends BC to the situation, where agents should take actions based on images, by utilizing a neural network architecture that contains CNNs and MLPs. For IRL, P-DAC in the concurrent work is employed to serve as the baseline because it demonstrates state-of-the-art performance.

\paragraph{Baselines for visual observations} We choose P-SIL and P-DAC in \citet{anonymous2022imitation} as counterparts for visual observations. Note that P-SIL and P-DAC are trained with expert data $\tau_E=\{(\bm{o})\}_0^N$, which slightly differs from the LfO setting. Hence, we conduct experiments to investigate the performance of OPIfVI with both $\tau_E=\{(\bm{o},\bm{o'})\}_0^N$ and $\tau_E=\{(\bm{o})\}_0^N$. To distinguish, we denote them as OPIfVI and OPIfVI-s, respectively. 

\begin{figure*}[!htbp]
\centering
    \subfigure[Visual demonstrations]{
    \includegraphics[width=1.0\textwidth]{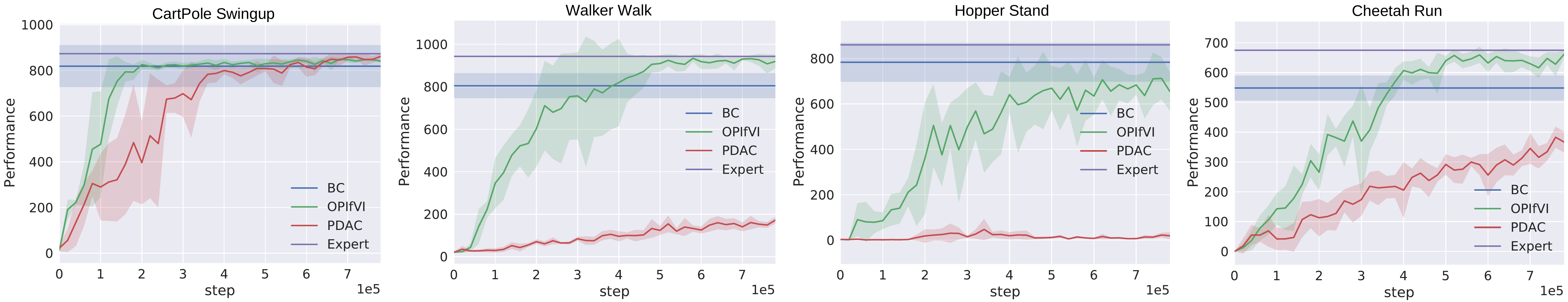}
    }
    \subfigure[Visual observations]{
    \includegraphics[width=1.0\textwidth]{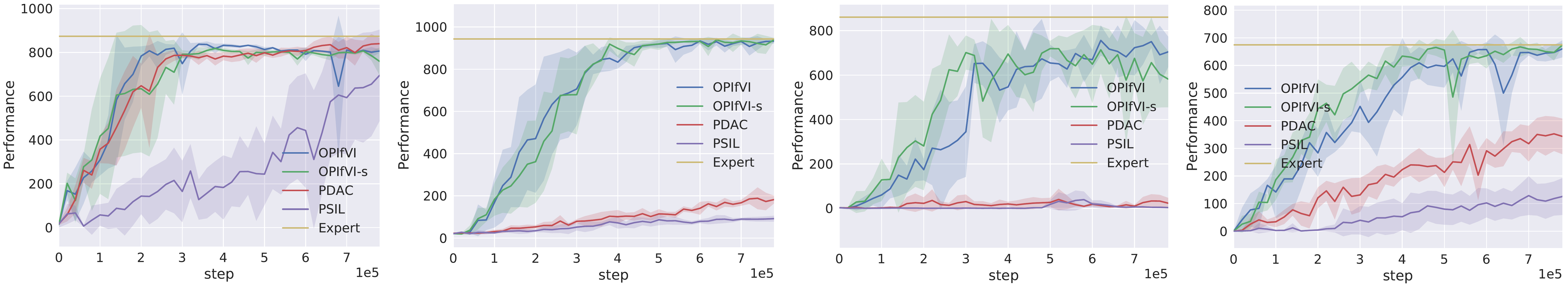}
    }
\caption{Performance of OPIfVI compared to other baselines on DeepMind Control tasks. Performance is measured with episode cumulative rewards, which is averaged across 5 random seeds, and the x-axis is the number of interactions with the environment.}
\label{learningcurve}
\end{figure*}

\subsection{Results}
We conduct experiments of OPIfVI with both visual demonstrations and visual observations. Besides, baselines corresponding to the two spectra are implemented and evaluated. The learning curves are shown in Figure \ref{learningcurve}. It is clear from the learning curves that: (1) OPIfVI is able to primely replicate expert behaviors and achieve expert-level performance no matter visual demonstrations and observations are provided, while the other baselines fail to perform as similar as experts with same training steps; (2) OPIfVI noticeably outperforms the baselines regarding both final performance and data-efficiency. For example, at 700k steps in Walker Walk, OPIfVI achieves about $\bm{6.6}$× and \textbf{5.5}× higher scores compared to PDAC in visual demonstrations and  visual observations, respectively.

\subsection{Ablation Studies} \label{subsection:ablationstudy}
\begin{figure*}[!htbp]
\centering
    \subfigure[Visual demonstrations]{
    \includegraphics[width=1.0\textwidth]{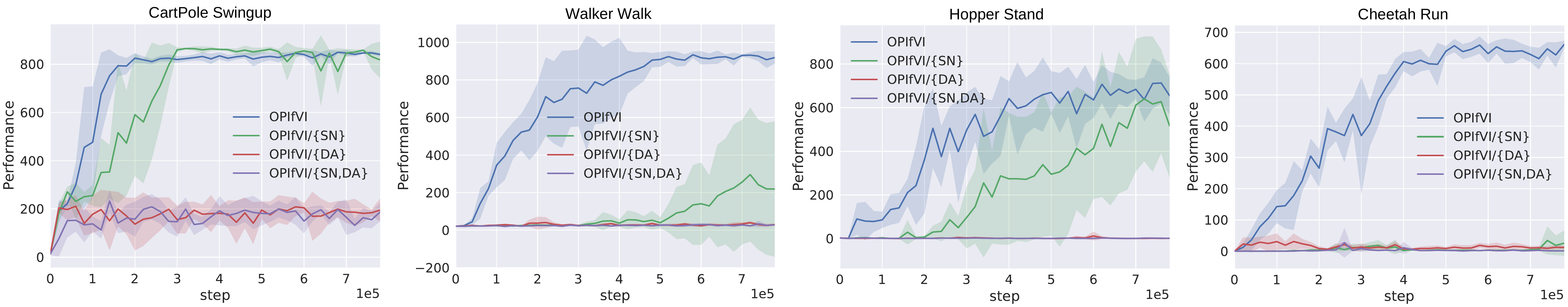}
    }
    \subfigure[Visual observations]{
    \includegraphics[width=1.0\textwidth]{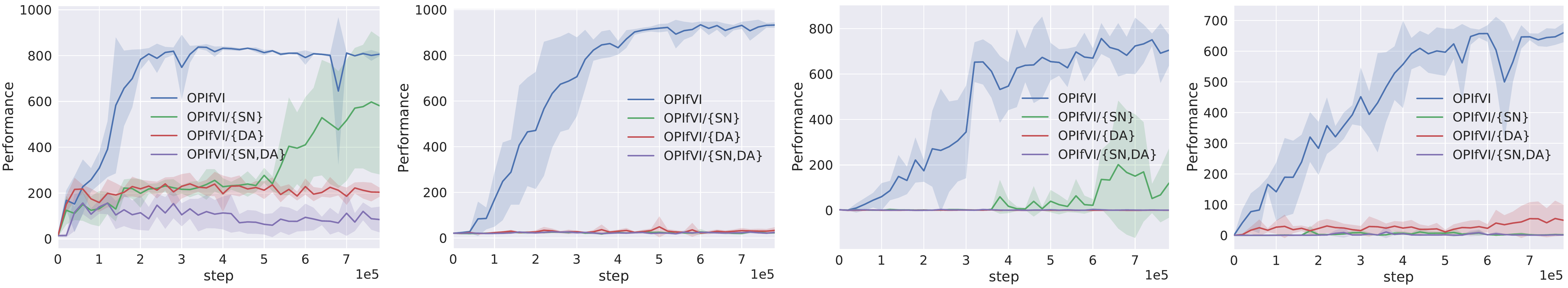}
    }
\caption{Ablation study of spectral normalization (SN) and data augmentation (DA) in OPIfVI. We use OPIfVI to represent the integrated framework, OPIfVI/{$x$} to denote that the framework without modification $x$. $x$ could be SN, DA, or any union of these two modifications.}
\label{ablationcurveDASN}
\end{figure*}

First, we study the impacts of spectral normalization and data augmentation in OPIfVI. Concretely, we compare the performance of OPIfVI against its versions without spectral normalization and/or data augmentation, whose results are visualized in Figure \ref{ablationcurveDASN}. From Figure \ref{ablationcurveDASN}, we can see that OPIfVI even could not reproduce a satisfactory policy to mimic experts in most environments without either of them. For example, in CartPole Swingup, OPIfVI/DA only achieves about a quarter of the OPIfVI's performance with visual demonstrations. In more complex environments, only OPIfVI is able to achieve expert-level performance, which means that both spectral normalization and data augmentation play an important role in OPIfVI.

\begin{figure*}[!htbp]
\centering
    \subfigure[Visual demonstrations]{
    \includegraphics[width=1.0\textwidth]{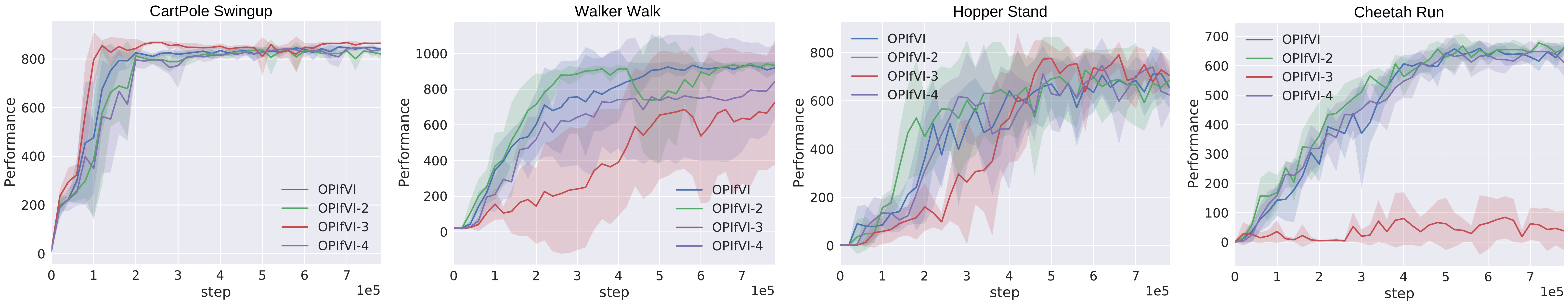}
    }
    \subfigure[Visual observations]{
    \includegraphics[width=1.0\textwidth]{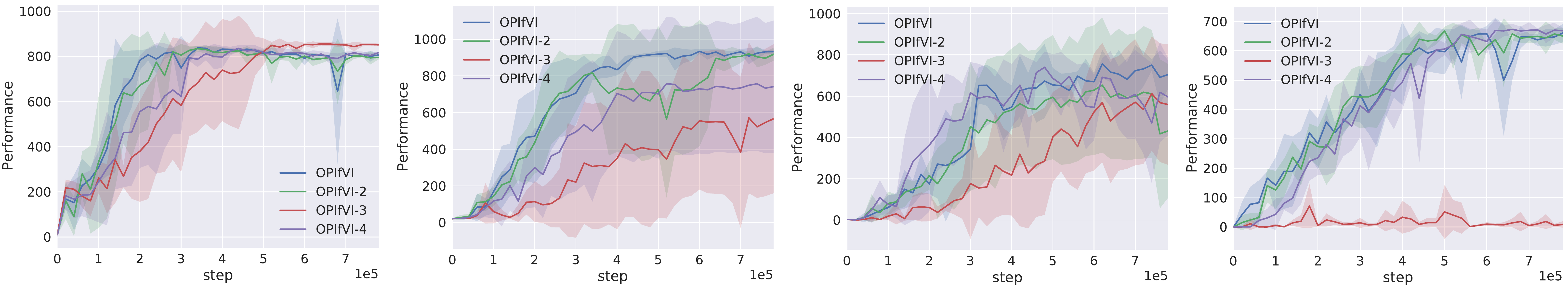}
    }
\caption{Ablation study of encoder structure in OPIfVI.}
\label{ablationcurveencoder}
\end{figure*}

Second, we conduct additional experiments to validate the encoder training structure. We test four cases: 1) the discriminator maintains a separate encoder and trains it with discriminator losses from scratch (OPIfVI, the structure we adopt); 2) similar to 1) despite that the encoder is trained with both losses from the actor and critic (OPIfVI-2); 3) the discriminator owns an encoder that is shared from the Q-value network and this encoder is trained with only the loss of Q-value functions (OPIfVI-3); 4) the discriminator, actor, and critic possess an independent encoder, respectively, and separately trains their encoders (OPIfVI-4). From the experimental results in Figure \ref{ablationcurveencoder}, we can see that OPIfVI demonstrates excellent performance across different environments and tasks. On the contrary, the other encoder structures could be unstable and perform poorly, especially on Walker Walk and Hopper Stand tasks with visual observations.

\section{Conclusion}\label{sec:conclusion}
\label{sec:conclusion}
In this paper, we present an imitation learning algorithm, OPIfVI, which can efficiently and effectively learn from visual inputs. OPIfVI works in an off-policy manner with stability enhanced with spectral normalization, improving learning efficiency. In addition, to deal with visual inputs, we adopt data augmentation and design a specific architecture to train the encoder. These two techniques help agents to better identify meaningful features in visual inputs, thus empowering agents to take correct actions. Compared to previous baselines, OPIfVI outperforms them regarding both data-efficiency and final performance.

\bibliography{iclr2022_conference}
\bibliographystyle{iclr2022_conference}

\appendix
\section{Environments and Expert Data}
\subsection{Environments and Specifications}
\begin{figure}[h]
  \centering
      \subfigure[CartPole Swingup]{
         \includegraphics[width=0.2\textwidth]{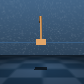}
      }
      \subfigure[Walker Walk]{
         \includegraphics[width=0.2\textwidth]{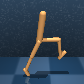}
      }
      \subfigure[Hopper Stand]{
         \includegraphics[width=0.2\textwidth]{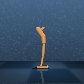}
      }
      \subfigure[Cheetah Run]{
         \includegraphics[width=0.2\textwidth]{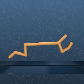}
      }
  \caption{Screenshots of the tasks in DeepMind Control Suite.}
  \label{fig:screenshotofenv}
\end{figure}

We choose DeepMind Control Suite \citep{tassa2018deepmind} to benchmark the performance of IL algorithms, which is widely adopted in previous works \citep{laskin2020reinforcement,anonymous2022imitation}. DeepMind Control Suite provides various image-based continuous control tasks, which is suitable for comprehensive evaluations in the setting of ILfVI. Four tasks with different complexities are employed in our experiments, \emph{i.e.}, CartPole Swingup, Walker Walk, Hopper Stand, and Cheetah Run, which are shown in Figure \ref{fig:screenshotofenv}. In our experiments, the agent take action according to three consecutive RGB images, and the height and width of images are set to 84 pixels. These configurations are coherent with \citet{yarats2020image}. Specifications of the tested tasks are presented in Table~\ref{dmcontroltask}. 

\begin{table*}[h!]
\centering
    \caption{Specifications of the DeepMind Control Suite tasks.}
\label{dmcontroltask}
\centering
\begin{tabular}{l c c c c}
  \hline
  Environment     & State Space  & Image Space   & Action Space & Max-Step\\
  \hline
  CartPole Swingup & 4 & $84 \times 84 \times 3$ & 4  & 1000\\
  Walker Walk     & 18 & $84 \times 84 \times 3$ & 6  &  1000\\
  Hopper Stand     & 14 & $84 \times 84 \times 3$ &  4 &  1000 \\
  Cheetah Run     & 18 & $84 \times 84 \times 3$ & 6 &  1000 \\
  \hline
\end{tabular}
\end{table*}

\subsection{Expert Data}
Here, we give more details on how we generate expert data in the experiments. As mentioned before, we use the algorithm DrQ \citep{yarats2020image} to train experts with default configurations. The action repeat for the four environments is set to 4. With an action repeat of 4, the episode length will be 250. We stack three consecutive frames together to construct $\bm{o}_t$. As a result, agents take actions according to $\bm{o}_t$ whose dimension is $84\times84\times9$ (channel last). Once we obtain trained expert policies, we can execute a policy of them in an environment and record the data. The recorded data are used as expert data. The expert data for visual demonstrations and observations are recorded as $\tau_E=\{(\bm{o},a)\}_0^N$ and $\tau_E=\{(\bm{o},\bm{o'})\}_0^N$, respectively. In particular, we also consider a special kind of visual observations that $\tau_E=\{(\bm{o})\}_0^N$. For every environment, we sample 20 expert trajectories, \emph{i.e.}, we store 5,000 state-action pairs or state-state pairs. The performance of expert data is listed in Table~\ref{expertdataperformance}.

\begin{table}[h]
\centering
\caption{Performance of expert data.}
\label{expertdataperformance}
\centering
\begin{tabular}{l c c c}
  \hline
  Environment     & Expert  \\
  \hline
  CartPole Swingup & 873.8 $\pm$ 1.5 \\
  Walker Walk     & 943.1 $\pm$ 22.4 \\
  Hopper Stand     & 860.7 $\pm$ 48.6\\
  Cheetah Run     & 675.0$\pm$ 30.9 \\
  \hline
\end{tabular}
\end{table}

\section{Implementation Details}
Our algorithm OPIfVI is implemented based on two open source codes, DrQ \citep{yarats2020image} and OpenAI Baselines \citep{dhariwal2017baselines}. The generator in OPIfVI is highly based on DrQ despite not adopting the augmentation for Q-value functions. The discriminator owns an identical encoder structure as the generator, and it is spectral normalized. The counterparts that we compare to, P-SIL and P-DAC, are the anonymous official implementation in \citet{anonymous2022imitation}. Since that the expert data in \citet{anonymous2022imitation} are not provided, we use the expert data constructed in above subsection instead. For fair comparisons, the action repeat is set to 4, and the batchsize is set to 128, while the other hyperparameters are the default for P-SIL and P-DAC. The hyperparameters for our experiments are presented in Table \ref{hyperparam}.   

\begin{table}[h]
  \caption{Hyperparameters in experiments.}
  \centering
  \begin{tabular}{ l|l}
  \hline
  Hyperparameters& Value\\
  \hline
  Environment parameters & \quad\\
  \qquad Image size & $84\times84\times3$ \\
  \qquad Action repeat & 4 \\
  \qquad Frame stack & 3 \\
  \hline
  Common parameters & \quad\\
  \qquad Activation & ReLU\\
  \qquad Batch size  & 128 \\
  \qquad Optimizer  & Adam \\
  \qquad Encoder feature dim  & 50  \\
  \qquad Actor update frequency & 2 \\
  \qquad Critic update frequency & 1 \\
  \qquad Discriminator update frequency & 1 \\
  \hline
  SAC parameters& \quad\\
  \qquad MLP network size  & (1024,1024)  \\
  \qquad Discount & 0.99\\
  \qquad Learning rate  & $1 \times 10^{-3}$ \\
  \qquad Initial temperature $\alpha$  & 0.1 \\
  \qquad Ployak & 0.01 \\
  \hline
  Discriminator parameters & \quad\\
  \qquad Learning rate & $1 \times 10^{-4}$ \\
  \qquad MLP network size  & (1024,1024)  \\
  \hline
  \end{tabular}
  \label{hyperparam}
\end{table}

\end{document}